\documentclass[conference]{IEEEtran}
\usepackage{amsmath,amsfonts}
\usepackage{algorithmic}
\usepackage{algorithm}
\usepackage{array}
\usepackage[caption=false,font=normalsize,labelfont=sf,textfont=sf]{subfig}
\usepackage{textcomp}
\usepackage{stfloats}
\usepackage{url}
\usepackage{verbatim}
\usepackage{graphicx}
\usepackage{cite}
\usepackage{multirow}
\usepackage{xcolor}
\usepackage{tikz,xcolor,hyperref}

\begin{document}

\title{Graph Based Traffic Analysis and Delay Prediction}

\author{
    \IEEEauthorblockN{Gabriele Borg}
    \IEEEauthorblockA{University of Malta\\
    gabriele.borg.18@um.edu.mt}
    \and
    \IEEEauthorblockN{Charlie Abela}
    \IEEEauthorblockA{University of Malta\\
    charlie.abela@um.edu.mt}
}

%\author{Gabriele Borg, Charlie Abela\orcidB{}%~\IEEEmembership{Member, IEEE}
        % <-this % stops a space 
%\thanks{(Corresponding author: Charlie Abela)
%Gabriele Borg and Charlie Abela are with the Department of Artificial Intelligence, Faculty of ICT, University of Malta, Malta (e-mail: gabriele.borg.18@um.edu.mt, charlie.abela@um.edu.mt.}% <-this % stops a space
%}

\markboth{}%
{Shell \MakeLowercase{\textit{et al.}}: A Sample Article Using IEEEtran.cls for IEEE Journals}


\maketitle

\begin{abstract}
This research is focused on traffic congestion in the small island of Malta which is the most densely populated country in the EU with about 1,672 inhabitants per square kilometre (4,331 inhabitants/sq mi). Furthermore, Malta has a rapid vehicle growth. Based on our research, the number of vehicles increased by around 11,000 in a little more than 6 months, which shows how important it is to have an accurate and comprehensive means of collecting data to tackle the issue of fluctuating traffic in Malta. In this paper, we first present the newly built comprehensive traffic dataset, called MalTra. This dataset includes realistic trips made by members of the public across the island over a period of 200 days. We then describe the methodology we adopted to generate syntactic data to complete our data set as much as possible. In our research, we consider both MalTra and the Q-Traffic dataset, which has been used in several other research studies. The statistical ARIMA model and two graph neural networks, the spatial temporal graph convolutional network (STGCN) and the diffusion convolutional recurrent network (DCRNN) were used to analyse and compare the results with existing research. From the evaluation, we found that the DCRNN model outperforms the STGCN with the former resulting in MAE of 3.98 (6.65 in the case of the latter) and a RMSE of 7.78 (against 12.73 of the latter).
\end{abstract}


\section{Introduction}
\IEEEPARstart{A}{s} the population and urbanisation increase, transport structures must also grow, resulting in important and complex infrastructures. However, transportation growth leads to problems that, in the case of densely populated areas, result in other problems such as increased air pollution and traffic congestion, especially during rush hours. Through the use of modern technology, traffic forecasting solutions are being thoroughly sought after in hopes of improving the efficiency of mass movement from one destination to another. This is being achieved through the use of different forms of sensors, such as street-level cameras and GPS data collected from public transport.
%\footnote{\url{https://nso.gov.mt/en/News_Releases/Documents/2021/0/News2021_122.pdf}}

The widespread adoption of mobile devices has enabled the continuous, implicit and instantaneous collection of vast amounts of data, ushering in a new era of data-driven insights. This phenomenon, often referred to as "big data," has significant implications for traffic analysis and prediction, as it allows for the collection of detailed spatio-temporal data at an unprecedented scale. According to Zhou et al. \cite{zhou2021}, the integration of data from mobile devices and sensors into intelligent transportation systems offers a unique opportunity to uncover hidden patterns and trends through advanced data analysis techniques. These patterns, which might not be apparent through traditional data collection methods, are crucial for developing more accurate and responsive traffic management strategies.

Using the vast amounts of data generated by mobile devices, researchers can address both known and unknown variables in traffic behaviour, leading to more robust and predictive models. As described by Li et al. (2018), analyzing large-scale traffic datasets enables the identification of previously unrecognized relationships and trends, thus enhancing the predictive power of traffic models [2]. This approach underscores the importance of big data analytics in transforming raw data into actionable insights for intelligent transportation systems.

The ubiquity of devices and their adoption across the population allows for a new phenomenon of collecting information implicitly, continuously, and instantly, from mobile devices. Furthermore, having such voluminous data stored in the palm of our hands, it is possible to discover unknown patterns and trends by analysing and processing such information. 

In this regard, the information collected in real time from mobile data users has been used to address the problem of traffic analysis \cite{saliba2018vehicular} in Malta, which is the most densely populated country in the EU with about 1,672 inhabitants per square kilometre and an ever increasing number of vehicles. The Maltese National Statistics Office (NSO) reported an average increase of 29 vehicles per day for 2021 \cite{nso-2021}.

Traffic prediction was investigated by \cite{cui2019traffic}, who evaluated the previous traffic states of a particular location over a period of time and considered these states as data to forecast the duration of the same respective location. Data-driven approaches are commonly adopted by researchers to address such a problem, thus requiring historical data for prediction.

Predicting traffic is however, complicated and is considered a nontrivial problem as it presents several challenges. Firstly, traffic flow \cite{respati2018traffic} is intricate as it is influenced by numerous variables, such as weather, road conditions, road network design, and human behaviour, which can rapidly and unpredictably change. Secondly, obtaining precise data \cite{golze2022spatial} can be difficult since traffic data collection methods can be limited and expensive, and crowd-sourced data may be unreliable. Additionally, traffic prediction models often require data from multiple sources that are challenging to integrate due to varying data formats and quality levels. 

The modelling of traffic prediction approaches can also be complicated, requiring significant computational resources, and selecting the best approach for a specific situation can be challenging \cite{zhang2021multiple}. Lastly, traffic prediction is time sensitive \cite{zhaoxia021realtime}, requiring models to be frequently updated and capable of adapting to real-time changes, which can be an arduous task. Overall, predicting traffic is a complex and challenging task that requires technical expertise and innovative approaches to overcome its challenges.

A comprehensive study conducted by Wu et al. \cite{wu2020comprehensive} highlights the effectiveness of Graph Neural Networks (GNNs) in various applications. The study demonstrates that GNNs can be applied to road structures by representing roads as nodes and their connections as edges. By doing so, GNNs can effectively address traffic problems by capturing spatial dependencies within geographical data.

Given the high volume of transportation and frequent traffic congestion in Malta, the current research aims to investigate this problem  , which is exacerbated by the size of the island and the number of vehicles. We describe how we collected and built the MalTra dataset that consists of travel information provided by more than 65 participants and how we used this dataset together with other similar datasets to analyse and evaluate recurring patterns in the current traffic scenario on Maltese roads using different GNN models.

\section{Motivations and Challenges}
This research highlights traffic-related challenges in Malta, which can be addressed using geolocation data in traffic management systems, directly or indirectly alleviating congestion. The focus of the research is on traffic patterns in general, followed by investigating the local Maltese sphere to predict the time spent on roads during a user's journey from their origin to the final destination.

The primary objective of this research is to develop a model capable of extracting insights from historical data (both spatial and temporal) to accurately predict the duration of travel on specific road segments or sensor points. To this end, various traffic patterns will be analysed, and these insights will be applied to the locally constructed dataset, facilitating the generation of precise traffic flow predictions. This research explores the hypothesis that graph-based neural networks can effectively predict the duration of traffic flow in Malta, offering a robust solution to the island’s traffic challenges.

To achieve this objective, several key goals have been outlined and will be rigorously evaluated to ensure the overall effectiveness of the model.
\begin{enumerate}
    \item Conduct an extensive review of existing methodologies for collecting traffic flow data at specific locations and times. This phase involves the analysis of various international traffic datasets, with the goal of constructing a bespoke local dataset suitable for input into predictive traffic models.
    \item Develop and train machine learning models for traffic prediction based on existing research \cite{wu2020comprehensive, jiang2022} and employ established datasets such as METR-LA, Q-Traffic, and MalTra (described in Section \ref{datasets}).
    \item Evaluate and compare the model predictions using standard metrics such as the mean absolute error (MAE), the mean absolute percentage error (MAPE), and the root mean squared error (RMSE) (discussed in Section \ref{metrics}, following the practices established in the literature. 
\end{enumerate}
This approach seeks to bridge the gap between local traffic dynamics and predictive modeling, providing a scalable, data-driven solution to improve traffic forecasting and management on the Maltese islands.

The rest of this paper is structured as follows: we begin with a comprehensive examination of the traffic data architecture, emphasising the intricacies of spatial-temporal data and its significance in transportation systems. Next, we provide an analysis of the current transportation framework in Malta, contextualising the study within local traffic dynamics. The discussion then shifts to the theoretical foundation of graph theory, which is essential for understanding the application of Graph Neural Networks (GNNs) in traffic flow prediction. A detailed review of related systems and case studies follows, providing critical insights and facilitating a comparative analysis of various datasets used in traffic modelling. The methodology section elaborates on data preprocessing techniques, particularly methods for addressing incomplete datasets and irregular time intervals. Finally, we present the experimental framework, including the model architectures, performance evaluations, and results, concluding with an analysis of the model's impact on traffic prediction in Malta and the broader implications for intelligent transportation systems.

\section{Related Work}
Before tackling traffic prediction problems, it is crucial to have a comprehensive understanding of the type of data with which we are working. By understanding the data type, we can identify relevant features and trends that can help us develop the local custom-built dataset.
 
\subsection{Spatio-Temporal Data}
\begin{figure}[t]
         \centering
         \includegraphics[scale = 0.25]{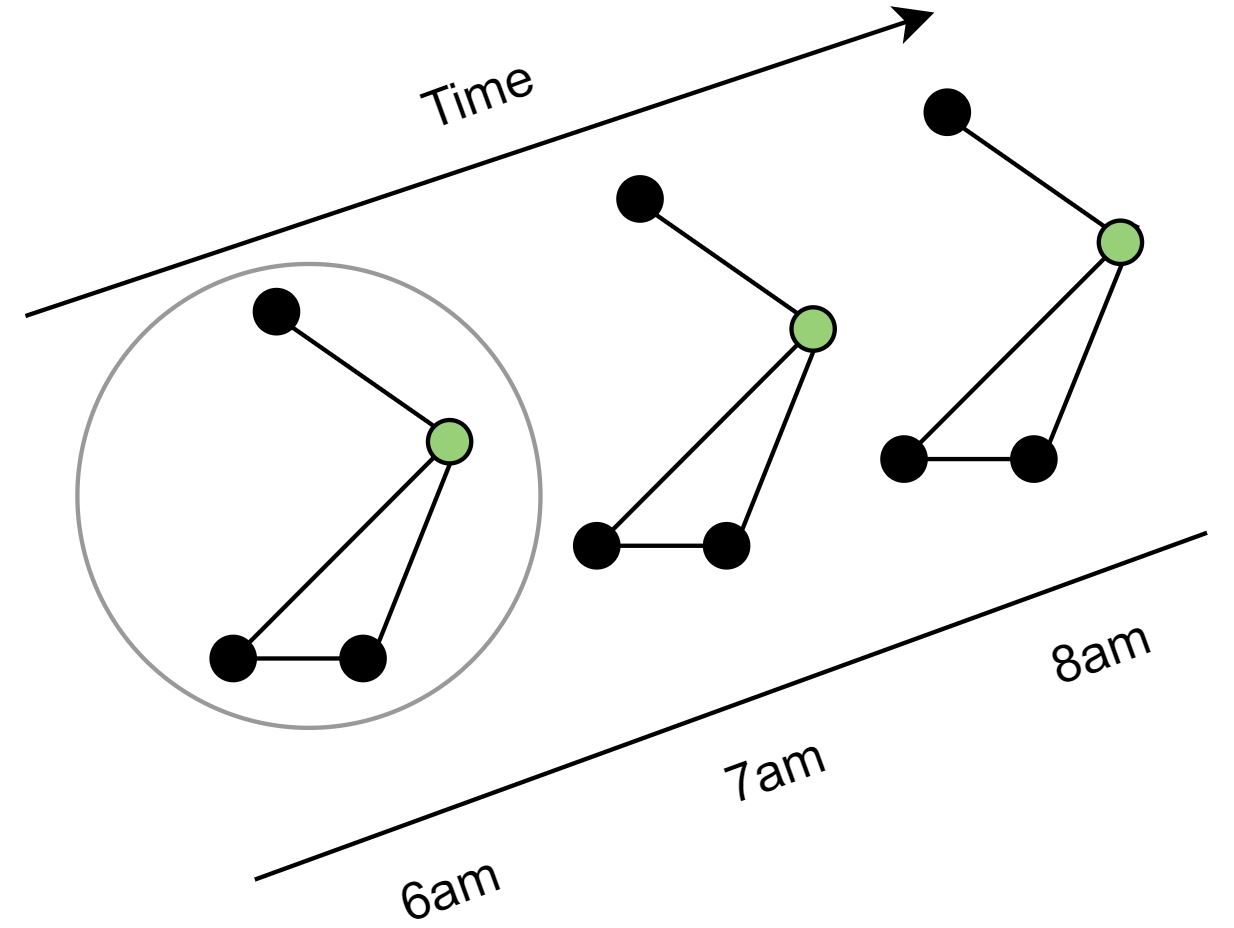}
         \caption{Spatio-Temporal node representation over a series of time}
         \label{fig:timeline}
    \end{figure}

 Spatio-temporal data refers to data that captures the spatial (space) and temporal (time) characteristics of traffic flow patterns in a space-time matrix.  Special dependencies refer to the road network's topological infrastructure while Temporal dependencies are the dynamic changes in traffic state over time while taking past time conditions into account. This can be visually represented in Figure \ref{fig:timeline}. Typically, the traffic flow or duration spent is examined at 5-minute intervals \cite{rico2021graph}. As a result, it can be inferred that each node corresponds to a dictionary of time intervals with the duration spent as its value. This information can then be utilized to generate a node-duration matrix, where each row corresponds to a node identifier and each column represents a 5-minute time step.
 
\subsection{Transport Malta}
Transport Malta (TM) is the governmental entity which is responsible for traffic management in Malta. Throughout recent years Transport Malta has been investing in various technologies and techniques to improve traffic situations within the island \cite{masterPlan}.
This transition to a more Intelligent Traffic System presents opportunities for Malta to monitor and analyse traffic patterns during the day, by collecting data from multiple sources, such as CCTVs, sensors, and Public Transport GPS data. 
However, Transport Malta can opt for alternative ways to collect data, even through the use of Big Data, just as MalTra was constructed.

\subsection{Graph Basics}
During the analysis of traffic forecasting, it has been observed that Graph Neural Networks (GNNs) are a suitable method for tackling such data. GNNs are at the forefront of modern deep learning research \cite{wu2020comprehensive}, and they excel in handling non-Euclidean data structures, such as road networks, which can be modelled as a graph. Non-Euclidean data refers to data that lacks an inherent structure.
A graph formulation of such an instance can be represented as a graph \textit{G}, where, 
     \begin{equation} \label{eq:graph-basic}
        G=(V,E,W)
    \end{equation}
    \textit{V} is the node-set (i.e.roads), \textit{E} the edge set, and \textit{W} is a weighted matrix (such as adjacency/feature matrix) \cite{rico2021graph}. An Adjacency matrix is a way of representing
    the edges between the nodes in a matrix structure of 1's and 0's (booleans), such that 1 denotes a connection (a direct path between the two nodes) and 0 denotes otherwise. Furthermore, a feature matrix reveals additional information/characteristics linked to the nodes themselves. This is usually in the form of a matrix \textit{F} with a dimension of \textit{n X f}, where \textit{n} is the number of nodes and \textit{f} is the number of features, with a representation of $X_t^i$, denoting the features of node \textit{i} at a given time \textit{t}.

\subsection{GNNs Models}
When reviewing literature about different types of GNNs currently being used for traffic prediction, two models stand out from others and are often used and compared when trying to treat such a problem. These are also known for being amongst the early models that were first developed \cite{rico2021graph,wang2018graph,li2017diffusion}, namely, Spatio-Temporal Graph Convolutional Networks (STGCN) and Diffusion Convolutional Recurrent NeuralNetworks (DCRNN).

In Wang et al. \cite{wang2018graph} the authors use STGCNs for traffic speed predictions at the sensor locations. This model combines GNNs and 1D-CNNs where they operate alternatively among the layers, where the Graph Convolutional Network (GCN) is responsible for capturing the spatial dependency (local neighbourhood of nodes) while the CNN is responsible for the temporal dependency (time data).
	
On a contrary note, Li et al. \cite{li2017diffusion}, explains how DCRNNs are different from STGCNs in certain ways. DCRNNs make use of RNNs rather than the 1D-CNN which is responsible for time dependencies, as well as spatial-based GCN instead of spectral approach (eigen approach) while incorporating the GCN inside a GRU. DCRNN rely on the information which is passed from one node to another also using an encoder-decoder system to predict future time steps. 

\subsection{Similar System}
    
Given that this research targets the local sphere we delve into similar works addressing Maltese traffic. Saliba et al. \cite{saliba2018vehicular}, presents a highly similar ideology to the proposed project. The novel approach includes a mixture of data mining followed but machine learning techniques, where the scope of this research was to investigate whether it is possible to predict vehicular traffic patterns based on mobile data. Although both the current research and this paper make use of mobile data, the projects differ in the method through which their data was obtained, as data in the latter paper is gathered from a telecommunications provider based in Malta. This data was used to generate an origin-destination (OD) matrix corresponding to the two most visited locations by a user. The authors propose a clustering approach to extract these two main locations, specifically the DBSCAN (density-based spatial clustering of applications) clustering technique. This is done by grouping the geographical coordinates based on a 500-meter radial coverage and checking which two clusters had the most activity going on. A spatial binning representation is then applied to the data, to smooth and reduce noise from the data, allowing important patterns to stand out, followed by a Multilayer Perceptron Classifier (MLPC) predictive model based on a neural network trained on grid nodes traffic levels. The findings obtained show a significant performance in predicting traffic flow demand within a location rather than the traffic flow rate itself.

Another research conducted by by Pace \cite{pace2017investigating}, also aims to target the traffic problem in Malta. For this investigation, the authors decided to gather a dataset of vehicle counts through the use of live-feed footage from an online webcam provider, Skylinewebcams \cite{skylinewebcams},
for a duration of seven days, during peak hours from 06:00hrs to 09:00hrs for six busy locations in Malta. These included, Marsa-Hamrun Bypass, Triq Mikiel Anton Vassalli in Kappara, Mriehel Bypass, Triq L-Imdina in Attard,  Msida Junction and St. Joseph High Street in Hamrun (the name of these locations is in line with those found in Google maps). However, due to missing data  the authors decided to re-dimensioned their research and focus only on three out of the six locations mentioned.
  
\subsection{Existing Datasets} \label{datasets}
    
When it comes to accuracy, models are only as good as the data they are fed. Therefore, this section investigated various datasets relating to the traffic scene highlighting any requirements that must be met when compiling the local dataset, referring to Objective \textbf{O1}.
Data of this kind is collected and sourced in a number of different ways by making use of Big Data \cite{zhu2018big, djedouboum2018big}, which refers to the collection of information from various devices and  sources.
     
Four datasets have been selected to model MalTra on their structure, namely Metr-La, Pems-Bay, Seattle-Loop-Data and Q-Traffic:

\begin{enumerate}
   \item \textbf{METR-LA:}
   {METR-LA (Los Angeles) \cite{jagadish2014big}
%   \footnote{https://paperswithcode.com/sota/traffic-prediction-on-metr-la}
   is one of the datasets that is frequently featured in traffic prediction studies. The METRA-LA dataset contains traffic speed data collected from over 200 loop sensors across the Los Angeles County highway. This collection has over six million traffic points ranging over a period of 4 months. Various papers \cite{yu2017spatio,zhang2020spatio,cai2020traffic,wang2020traffic,chen2001freeway} make use of these datasets and aggregate the records into 5-minute intervals resulting in twelve observations per hour, and 288 observations in a single day for more than 207 sensors.}

   \item \textbf{PEMS-BAY:} {Similarly, the PEMS-BAY dataset \cite{pemsbaypaper} shares the same structure and is collected by a transport agency in California. Containing over 325 sensors in the Bay Area and collected over 6 months during 2017, PEMS-BAY also aggregates traffic speed readings into five-minute windows.}
   
   \item \textbf{Seattle-Loop-Data:} {This data is collected through inductive detectors across four freeways in Seattle, namely, I-5, I-405, I-90, and SR-520. These loop detectors collect spatiotemporal data regarding the speed of traffic within the respective freeways with intervals of five minutes apart for each record. With over 323 sensors, traffic state data of more than 1000 roadway segments is recorded for the year 2015, with an updated version in 2021.}
   
   \item \textbf{Q-Traffic:} {This is a large-scale traffic dataset which contains 3 sub-datasets, namely traffic speed, road networks and direction query sub-dataset. The sub-dataset which is most appealing to this research, and other papers \cite{liao2018deep,yin2021deep}, is the 'traffic speed dataset' which  contains 15,073 road segments with 15-minute time intervals over 61 days. Each record is represented as \textit{road\_segment\_id}, \textit{time\_stamp ([0, 5856])} and \textit{traffic\_speed (km/hr)}. This is collected in Beijing, based on Baidu maps \cite{bidumaps}}

   \item \textbf{TaxiBJ:} {This dataset is sourced from over 34,000 taxis' GPS routes which are situated in the city of Beijing \cite{zhang2017deep}. Traffic data volumes are collected and aggregated into 30-minute intervals, in a bi-directional manner (In and Out) and are compiled over the course of four months (1st March to 30th June 2015). The raw dataset is around 3GB and each record is made up of a \textit{taxi\_id\_number}, \textit{time (Unix)} and \textit{traffic\_speed (km/hr)}, \textit{longitude}, \textit{latitude} and a \textit{state} flag denoting whether the taxi is in service or has no passengers. This also included a 'weather condition' value categorized into five groups, namely, sunny, cloudy, rainy, storm and dusty.}

\end{enumerate}

    \subsection{MalTra Dataset} The Malta Traffic dataset is a custom-built collection sourced from multiple local participants, consisting of over \textbf{752 trips}, across \textbf{41 localities} with \textbf{602 unique roads}. These records were collected over the period of \textbf{200 days} (September 2021 - March 2022) \cite{gabriele2023}. 
    \begin{figure*}[h]
    \centering
        \includegraphics[width=0.8\textwidth]{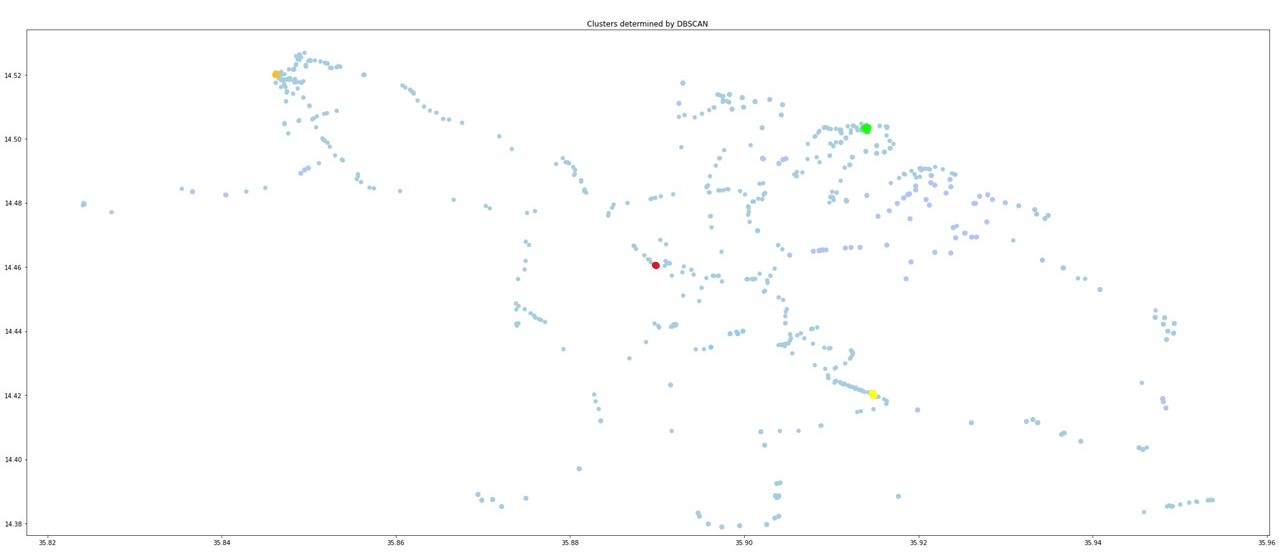}
        \caption{Data points of one user (light blue points) extracted from Google Timeline - an outline of Malta can be seen (including Saint Julian's, Valletta and the Three Cities) - own source}
        \label{fig:googleTimleinPoints}
    \end{figure*}
    
    This collection consists of data records gathered from Google Timeline, thus it is not  considered a traffic dataset but rather a collection of data records, as seen in Figure \ref{fig:googleTimleinPoints}. Google Timeline is a component of the Google Maps package that enables users to view their past travel routes and export them. However, the data collected by this feature is not deterministic and incomplete meaning that if the same trip is taken twice, the resulting set of points may not be identical and may fail to record every road travelled. To address these issues, a multi-step pipeline has been developed to filter the data obtained from Google Timeline and convert it into a traffic dataset for further analysis.
    
\begin{figure}[t]
        \centering
            \includegraphics[width=0.5\textwidth]{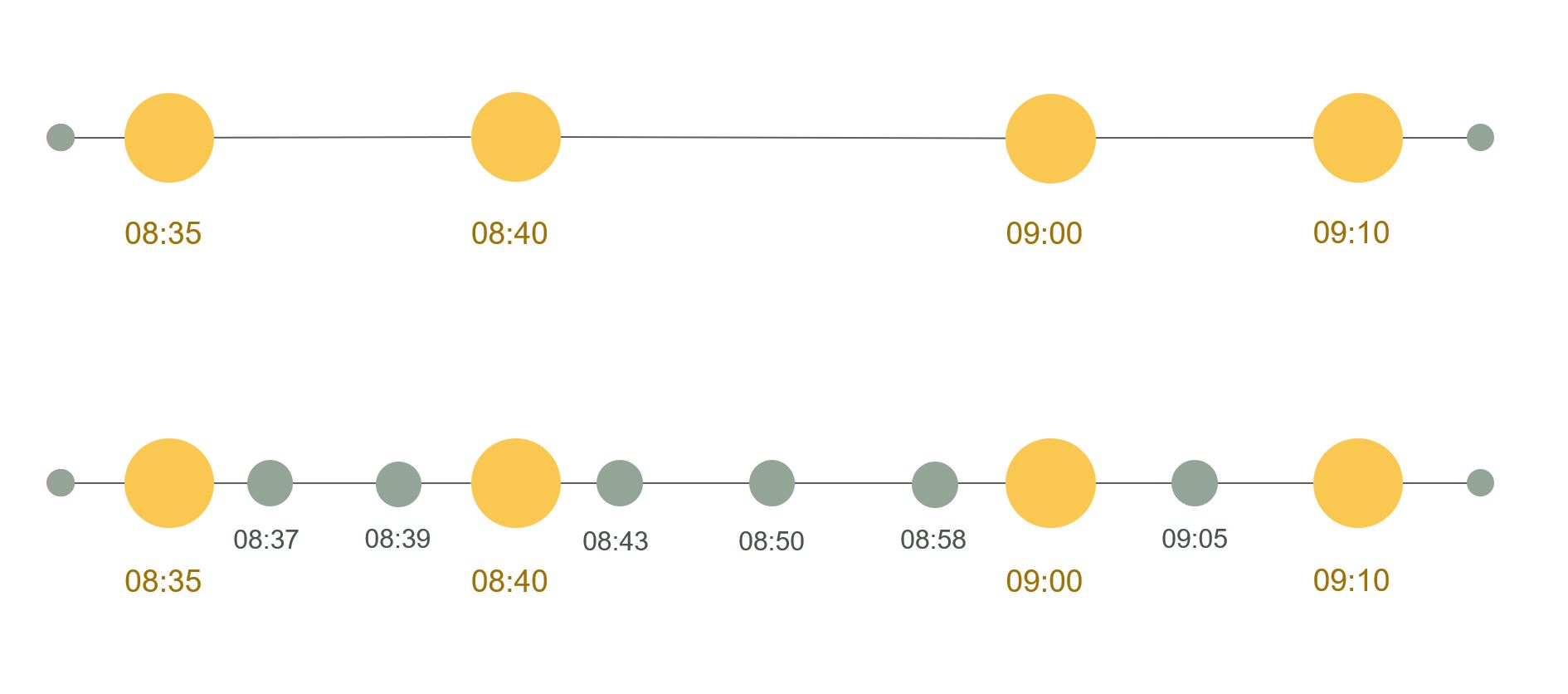}
            \caption{Generating new data points between the existing ones, referred to as 'checkpoints', which are obtained from Google Timeline.- own source}
        \label{fig:checkpoints}
    \end{figure}

    In our data extraction process, we checked various phone models, operating systems and app versions to ensure all the necessary data was uniform. We furthermore identify and parse the "Home" and "Work" locations, clustering the most active areas using DBSCAN (500-meter radius), and grouping data entries by road identifiers to ensure accurate duration times. However, we encountered missing roads, which we retrieved using Open Street Maps (OSM) API. Then for each trip we generated timestamps by splitting the parent timestamp proportionally based on the distance of each road and assigned a timestamp group property in five-minute intervals, see Figure \ref{fig:checkpoints}. Finally, we combined all extracted roads into one comprehensive with attributes such as road details and traffic flow values. Our process also involved many other additional checks to ensure good data, as described in a separate paper \cite{gabriele2023}.

 \hfill

Table \ref{table:datasets-table} is a summary of all the datasets discussed.
    
\begin{table}[ht]
    \centering
      \caption{Dataset Features }
        \begin{tabular}{llll}
        \hline \hline  \\ [0.5ex]
        Title  &  Sensors   &  Entries  & Intervals \\ [2ex]
        \hline \hline 
        \\
        METR-LA & 200  &  34,272  & 5 mins\\ [1ex]
        PEMS-BAY &  325  & 52,116 & 5 mins\\ [1ex]
        Seattle-Loop  & 323  &   3,210,736 & 5 mins\\ [1ex]
        Q-Traffic  & 15,073 &   265,967,808 & 5/10/15 mins\\ [1ex]
        TaxiBJ  & 500  &   12,205,120 & 30 mins \\ [1ex]
        MalTra   & 602  &   12,000 & 5 mins \\ 
        \\
        \hline
        
    \end{tabular}
    \label{table:datasets-table}
  
\end{table}

\section{Methodology}
This section gives a detailed description of the implementation that is carried out for this research and consists of generating synthetic data from the original collection of MalTra. This is a necessary step to address any missing entries in the dataset. We then analyse the data generated and train the STGCN and DCRNN models across three different datasets to obtain both a baseline and a comparison value. 
    
\subsection{Generating Synthetic Data}
As was indicated, Google Timeline was utilised to collect such data for MalTra; however, the primary intention of Google Timeline was not to carry out such a data collection task. As a result of this, Google Timeline is both nondeterministic and incomplete (as stated in Google's support page that it "shows an estimate of places that you may have been and routes that you may have taken" as well as that "occasionally, you may find mistakes on Timeline.") \cite{googleinfopage}. This means that if a user travels the same route twice, it is possible that it will output two different sets of points while being incomplete in the sense that not all roads within a particular route may be identified and recorded.

As a consequence of this, there are a few missing records in between journeys, resulting in an incomplete series of time interval records. Since the majority of users adhere to a daily schedule or routine during the course of the day, they are likely to pass through the same road around a similar time, thus leaving certain time intervals without a corresponding value, as seen in Table \ref{table:gaps}. To use MalTra for traffic forecasting, the first step to be done is to generate and populate any missing values with synthetic data.

    \begin{table}[t]
        \centering
        \caption{Time series containing missing records}
        
        \begin{tabular}{c c c c c c c} 
                
        \hline \hline 
        &   &     \\
        Road Name         &    Duration & Time Interval \\
        &   &     \\
        \hline \hline 
        &   &      \\
        Triq Di\.{c}embru 13, Il-Marsa  & 115.37   & 7:45  \\
        &   &      \\
        Triq Di\.{c}embru 13, Il-Marsa  & 125.37   & 7:50  \\
        &   &      \\
        Triq Di\.{c}embru 13, Il-Marsa  &  - -  &  7:55 \\
        &   &          \\
        Triq Di\.{c}embru 13, Il-Marsa & 61.22   & 8:00 \\
        &   &        \\
        Triq Di\.{c}embru 13, Il-Marsa & 59.58    & 8:05  \\
        &   &         \\
        \hline
        \label{table:gaps}
    \end{tabular}
    \end{table}

       Our process consists of two steps: the first is to generate missing values for missing days, and the second is to generate missing values for certain time intervals.

 \begin{enumerate}
 
    \item \textbf{Missing Days Generation:} {The first step of this operation is the generation of missing records for specific days throughout specific time periods. In this particular scenario, we have a general sense of the amount of time spent in the corresponding location (sensor) based on our experiences from earlier days available throughout the same time period.
    For example, in MalTra, if there are 200 days' worth of records, there may be a few missing values for a certain interval, such as 07:10. These can be modelled based on the other data at 07:10 by taking the average of the set that has been randomly chosen. 
    
    Consider a scenario in which 20 per cent of the values are missing. In this scenario, the average of a random collection of records drawn from the remaining 80 per cent of the dataset is computed and kept for each of the missing values.
    
    This procedure is illustrated in Figure \ref{fig:daygeneration}. This involves calculating an average value based on a random sampling of data from the previous records of the same time interval records whenever a day is absent from the data set.
    
    \begin{figure}[ ]
        \centering
        \includegraphics[width=0.5\textwidth]{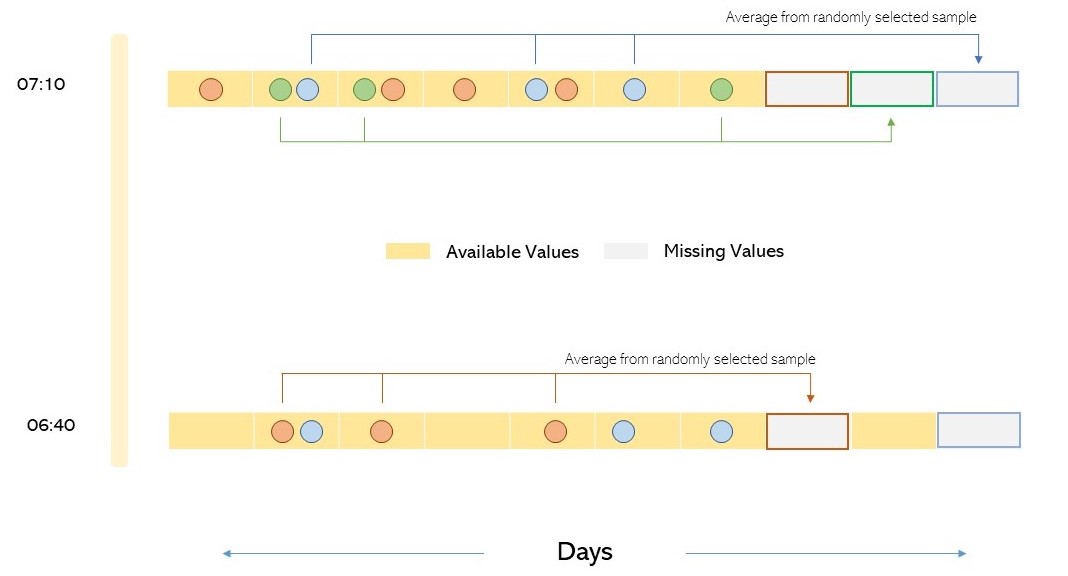}
        \caption{Calculating an average for missing day values using a random sample of a particular time interval from  prior records - own source}
        \label{fig:daygeneration}
    \end{figure}
    }

    \item \textbf{Time Interval Generation:}  {This involves the creation of new time interval values. This occurs during those time intervals in which no users passed from the said location (sensor), and hence, no records were ever made within that specific time range. Since the majority of users adhere to a predetermined plan and habit of a routine for the course of the day, it is likely that they will pass by the sensor at around the same point in time. This generation process is illustrated in Figure \ref{fig:timeintervalgeneration}.
    
    For each missing daily record, the adjacent time intervals are taken into consideration, and a linear progression is put into action to fill up the gaps.
    
    As suggested in \cite{patki2016}, the Synthetic Data Vault (SDV) was also considered during this process. SDV is an ecosystem of libraries for the generation of synthetic data that enables users to easily learn single-table, multi-table, and time-series datasets. These datasets can then be used to generate new synthetic data that has the same format and statistical properties as the original dataset. SDV permits the use of a variety of models, one of which is the Gaussian, which is created from a multivariate normal distribution over by employing the probability integral transform. Other models that can be constructed with SDV include Conditional modelling and Linear Regression.
    
    Since the intervals are only five minutes apart, the change in time is not significant, and as a result, the last option was selected.
    
    \begin{figure}[h]
        \centering
        \includegraphics[width=0.5\textwidth]{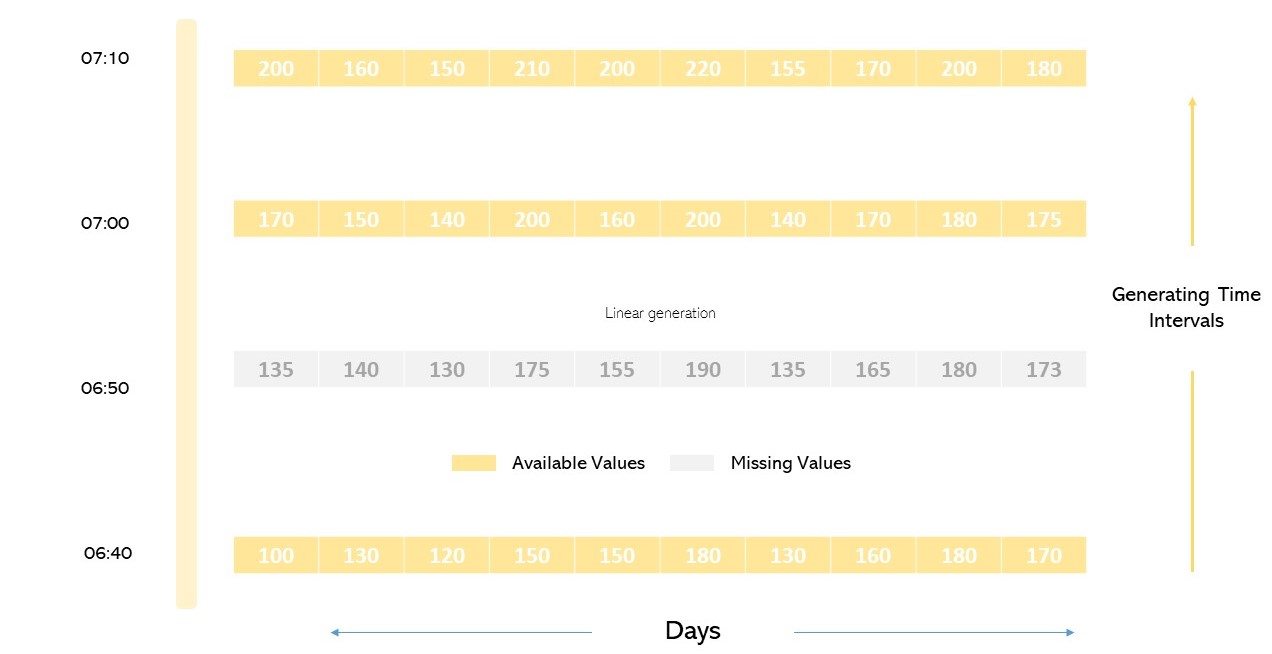}
        \caption{Neighboring time periods are considered and a linear progression is used for missing daily records - own source}
        \label{fig:timeintervalgeneration}
    \end{figure}
}
    
    \end{enumerate}
    
After this procedure is carried out, the final product is a comprehensive dataset where each time period across each day has a value that corresponds to it. Because of this, we are able to organise the data in the appropriate input format so that it is ready for training.

\subsection{Data Analysis}
We first applied the ARIMA model on MalTra.  The advantage of ARIMA is that it can address the stationary problems associated with time-series data. A time series is said to be stationary if there are no trending or seasonal impacts on it.

As suggested by \cite{li2017diffusion}, instead of feeding all the timestamps to the ARIMA Model, they focused on a period of one week, while the prediction was based upon the aggregated values of the former weeks. Thus the forecasted value for a particular day of the week, such as next Monday, is the average traffic speed from the previous four Mondays. However, when really analysing this approach, it was concluded that a single average speed value for a corresponding day is not really helpful at all.

Stating that on Monday the speed will be \(X\) and on Tuesday it will be \(Y\), is not that useful. On the contrary, having an average speed which corresponds to a particular hour would be more beneficial, for example, at 06:00 the speed is \(X\) but at, 07:00, the speed is \(2X\).

However, while going down this route we quickly realized that for every interval there is only one corresponding value, i.e. the average of that interval. This would obviously not work as a model can not base any prediction on one input value, thus this approach was not adopted. 

\begin{figure*}
        \includegraphics[width=1\textwidth,height=4cm]{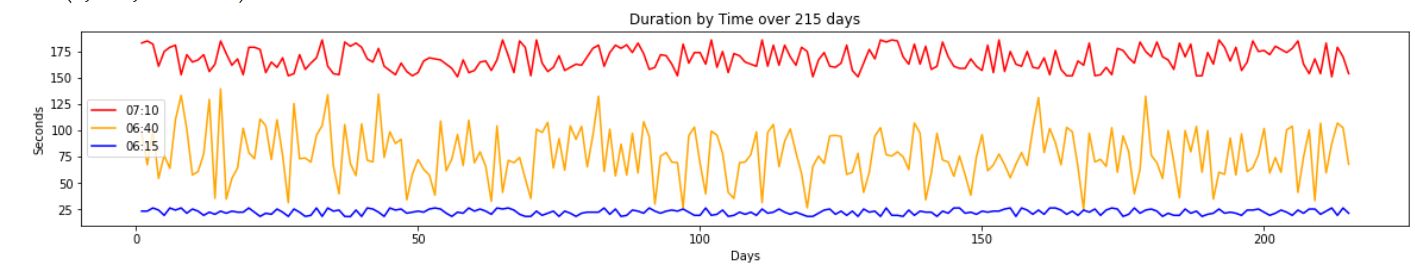}
        \caption{Difference in traffic duration by three-time interval groups (Floriana) - own source}
        \label{fig:3intervals}
    \end{figure*}

The alternative approach targeting individual time intervals themselves rather than the whole time series is to extract all the duration values recorded for a selected interval. In this way, a slice is extracted from the dataset with all the intervals and special focus is zoomed into one particular interval, thus being able to predict the next couple of records of hour \textit{X} based on all the previous hour \textit{X}, irrespective of the day.
A visual of this viewpoint can be seen in Figure \ref{fig:3intervals}, namely focusing on 06:15 (blue), 06:40 (orange) and 07:10 (red). 

It is interesting to see that at 06:15 the duration spent on the road is quite low, furthermore, it holds a small amplitude displacement, as it is fairly given that at that time there is no traffic, irrespective of the days or commuters' routine. 

At 06:40, the duration slightly increases, along with the amplitude. It is at this time that traffic starts to gather and to increase slowly, some days it gathers at quicker or at an earlier time while some days it is more relaxed and gathers at a later stage. 

Finally, at 07:10, traffic has reached its peak, with a small amplitude in the graph, indicating that irrespective of the day, traffic will likely be present in Floriana. Extracting the 07:10 timestamps as the historical data for this test, the Augmented Dickey-Fuller Test is now used to statistically check if the data is stationary or not. If the p-value is less than 0.05 then the data is stationary, else it is non-stationary and thus difference would need to be used. If the data is stationary then the difference is not needed and the ARIMA's d value is zero. More in-depth detail is given in the full paper.

\subsection{Models}

In the following section, we delve into the practical application of two effective models, namely the STGCN (Spatio-Temporal Graph Convolutional Network) and the DCRNN (Diffusion Convolutional Recurrent Neural Network), to address the complex challenges of traffic prediction within road networks. Our goal is to thoroughly evaluate how well these models forecast traffic statuses and flow using data from the actual world. Our research attempts to clarify their capabilities and offer perceptions about how well they can improve traffic control in Malta.

\subsubsection{The Application Of The STGCN Model}
In continuation of the work done by \cite{yu2017spatio}, our objective was to test this model on two selected datasets, namely, Q-Traffic and MalTra. The scope of this application was to determine whether or not it is possible to predict the traffic state in Malta, or rather, whether or not it is possible to predict the traffic flow based on crowd-sourced data. With the STGCN model, we essentially have two inputs to work with. The first component is a matrix \(X\) that contains node features; the shape of this matrix is determined by the batch size, the number of nodes in the graph, the number of features pertaining to each node, and, lastly, the number of previous timesteps that are to be taken into consideration. In addition, the other input is a matrix referred to as the Adjacency matrix \(A\), which defines the connections between the nodes in the graph.

These are then passed onto the temporal and spatial blocks, respectively. As a result, we are required to generate three files for MalTra listed below: the \textit{H5} data frame file (data file saved in the Hierarchical Data Format), the \textit{distances\_maltra.csv} file, and finally, the \textit{graph\_sensor\_ids.txt} file. These files are then used to generate an adjacency matrix, which the model will later use to create a graph. The \textit{distances\_maltra.csv} contains the connections between points as well as the respective distance for each link. To conduct this, we opted for Open Route Service \cite{openrouteservices}, an open-source interactive API. Finally, complete the final requirement, that's is, \textit{graph\_sensor \_ids.txt}. As the name suggests, this file contains a list of all the sensor identifiers found within MalTra.

\subsubsection{The Application Of The DCRNN Model}

In this section, we start implementing a DCRNN model to have a comparison set of results to those produced by the STGCN. The DCRNN shares a different structure, implementing bidirectional random walks to capture the spatial dependency among the nodes in the graph. We also employ an encoder-decoder architecture with scheduled sampling in order to capture the temporal dependency \cite{selvi2021diffusion}.  Historical time data serve as input for the encoder. The encoder then converts the sequence of historical time  data into a vector with a specified length. The decoder will then predict what the next timestamp output from the encoded vector will be. The use of recurrent neural networks is implemented in order to handle the temporal dependency. The stochastic gradient approach can be used to train the diffusion convolutional layer, which can then be used to learn the graph representation. Gated Recurrent Unit (GRU) is employed with a diffusion convolutional layer because it can accept long-term dependencies better. 

Prior to applying the datasets to the model, an input file structure is required for MalTra. The model requires three distinct files, namely, an H5 data frame file, the \textit{distances\_maltra.csv} file, and the \textit{graph\_sensor\_ids.txt} file.  In addition to these, it also requires two additional data files namely, the \textit{graph\_sensor\_locations.csv} and the adjacency matrix denoting the node connectivity, which can be found in the \textit{adj\_mx.pkl} file. 

\section{Results} 

In this section, we evaluate the models compiled in the methodology on the selected datasets. It is crucial to keep in mind that traffic in Malta is one of the most pressing issues currently persisting on the island. According to a new report published by the NSO \cite{nsocarsperday} an average of 54 motor vehicles per day were newly licensed during the final and first quarters of 2021 and 2022, respectively. This means that an addition of over 11,600 cars was introduced between the first and last day of the data collection process.
\subsection{Analysing traffic flows across various intervals}
Recall the data in Figure \ref{fig:3intervals}; in order to visually develop an intellectual comprehension of the data that is being analysed, we developed a box plot to visualize the dispersion and skewness of the data. Furthermore, we opted for the Anova model \cite{st1989analysis} which is a collection of statistical models used to analyze if the differences between the means of groups are statistically significant. In this case, we examine if there is a significant difference between traffic duration across different hours. this test resulted in a significant value lower than 0.05, which means that the null hypothesis is rejected, thus the results demonstrate that a significant difference does in fact exist, thus implying rush hours.

\begin{figure}[h]
    \centering
        \includegraphics[width=0.45\textwidth]{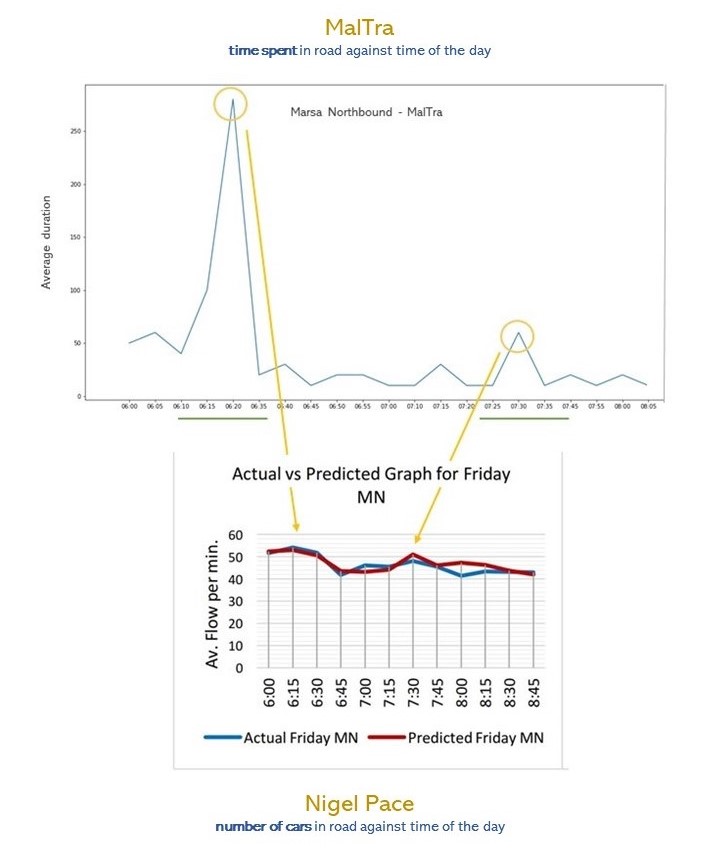}
        \caption{Marsa northbound traffic flow comparison between MalTra and Figure 4-19 in research by author Nigel Pace \cite{pace2017investigating}}
        \label{fig:comparison}
    \end{figure}

While analysing such data, we came across an interesting resemblance shared by two researchers, this paper and \cite{pace2017investigating}. In Figure \ref{fig:comparison} two graphs are displayed; the top one belongs to MalTra, while the other belongs to \cite{pace2017investigating}. Both of these indicate the flow of traffic in the Marsa region, more specifically in the Marsa Northbound lane, which is the lane that directs traffic from the southern side to the northern side of the island. When analysing both graphs in Figure \ref{fig:comparison}, it can be seen that the time series is dominated by two peaks, one at 06:15, which is the highest one, and the other one at 07:30. Even though different measurements are being utilised, such as the average automobiles per minute and the average duration spent in the road, both reveal that there has been an increase in the traffic flow. Although MalTra's graph possesses a greater amplitude difference, both of these peaks can be seen at the same timestamps across the two studies, respectively. The difference in amplitude displacement can be substantially justified due to the difference in the used y-axis metric. An addition of 15 cars appears to be a rather slight increase in \cite{pace2017investigating}'s graph; however, the same 15-car change translates to much more seconds in duration time spent within the respective road.

\subsection{Metrics} \label{metrics}
Various evaluation metrics have been seen being employed as an evaluation measure in a number of studies \cite{cui2019traffic,selvi2021diffusion,li2017diffusion,yu2017spatio, yin2021deep,guo2019attention,wu2019graph,jia2016traffic,zhou2020variational,chen2020dynamic,chen2020tssrgcn,wang2018graph,bbliaojqZhangKDD18deep,jiang2022}, namely, MAE (Mean Absolute Error), MAPE (Mean Absolute Percentage Error), and RMSE (Root Mean Square Error).

\begin{enumerate}
    \item \textbf{MAPE} {The mean average percentage error (MAPE) is a measurement that determines how accurate a forecast system is. This accuracy is expressed as a percentage, and it may be determined by taking the average absolute per cent inaccuracy for each time period and dividing it by the actual values.
        
     \begin{equation} 
         MAPE (y,\hat{y}) = \frac{1}{M}   \sum_{i=1}^{M} \frac{|y_{i}-\hat{y}_{i}|}{y_{i}} 
     \end{equation}
    }   
    \item \textbf{MAE} {The mean average error (MAE) calculates the average magnitude of errors in a group of forecasts without taking into account the direction in which the errors are directed. It determines how well continuous variables are measured. To put it another way, the MAE is the average calculated, across the verification sample, of the absolute values of the differences that exist between the forecast and the observation that corresponds to it.
       
    \begin{equation} 
        MAE(y,\hat{y}) = \frac{1}{M}   \sum_{i=1}^{M} |y_{i}-\hat{y}_{i}|
    \end{equation}
    }
    \item \textbf{RMSE} {The root mean square error (RMSE) measures the variance of the prediction errors. This error measures how far away from the regression line the data points are and evaluates how dispersed these residuals are. In other words, it reveals the degree to which the data is centred on the line of best fit. 
   
    \begin{equation} 
       RMSE (y,\hat{y}) = \sqrt{\frac{1}{M}  \sum_{i=1}^{M} (y_{i}-\hat{y}_{i})^{2}} 
    \end{equation}
    }
\end{enumerate}

\subsection{Results Discussion DCRNN - STGCN}

Table \ref{table:comaprison-results} presents the performance comparison of two traffic prediction models, STGCN and DCRNN, on two datasets, METR-LA and Q-Traffic, using three evaluation metrics: MAE, MAPE, and RMSE. 

Considering the first, the STGCN model obtained an MAE of 6.65, a MAPE of 28.26\%, and an RMSE of 12.73 for the METR-LA dataset and an MAE of 5.07, a MAPE of 15.21\%, and an RMSE of 6.8 for the Q-Traffic dataset. Whilst the DCRNN achieved an MAE of 3.98, MAPE of 12.31\%, and RMSE of 7.78 for the METR-LA dataset and an MAE of 3.79, MAPE of 9.81\%, and RMSE of 4.22 for the Q-Traffic dataset. These findings suggest that DCRNN is a stronger model for traffic prediction than STGCN. 

\begin{table}[h]
    \caption{Performance baseline comparison between two large-scale datasets, METR-LA and Q-Traffic }
    \begin{tabular}{|c|| c| c | c || c |c |c| } 
            \hline 
             &  \multicolumn{3}{|c||}{METR-LA} &  \multicolumn{3}{c|}{Q-Traffic} 
            \\  \cline{2-7}
            \multirow{2}{*}{Model}  & & & & & & \\
            & MAE & MAPE &  RMSE & MAE & MAPE &  RMSE \\        
            \hline\hline 
            & & & & & & \\
            STGCN & 6.65  & 28.26 & 12.73  & 5.07  & 15.21 & 6.89    \\ [1ex] 
            DCRNN & 3.98  & 12.31 & 7.78  & 3.79  & 9.81 & 4.22   \\[1ex] 
            \hline
    \end{tabular}
            \label{table:comaprison-results}
    
\end{table}

\subsection{Applying the MalTra Dataset}

Following an analysis of the two datasets using the two distinct models, the results conclude that DCRNN is the more competitive model when it comes to traffic prediction, leading the DCRNN to be obtaining better outcomes. As a consequence of this, the last stage of this evaluation will consist of evaluating the MalTra on the DCRNN model.
    
Due to the fact that Malta is a significantly smaller island compared to both Los Angeles or Beijing, with a total land area of 316 km\(^2\), 1,299 km\(^2\), and 16,411 km\(^2\) respectively, the island of Malta does not include any comparable motorways. As a result of this rationale, the six sensor locations with the most data, during the collection period among the participants, are taken into account. These are \textbf{Lucija, Marsa, Mosta, Attard, Msida and Furjana}. Table \ref{table:selected-six-sensors} shows the density and the percentage coverage of the dataset.

      \begin{table}[h]
            \caption{Selected sensor locations with the most records from MalTra - own source}
                \begin{tabular}{|ccc||ccc|ccc|} 
                \hline  
                & \textbf{Locality (Sensor)} & & & \textbf{Records} & & & \textbf{Percentage} & \\  
                \hline \hline 
                & Marsa & & & 2,809 & & &  23.18 \% & \\ 
                & Msida & & & 1,823 & & & 15.04 \% & \\ 
                & Mosta & & & 1,754 & & & 14.48 \% & \\ 
                & Furjana & & & 821 & & & 6.78 \% & \\  
                & Attard & & & 558  & & & 4.61 \% & \\ 
                & Lucija & & & 459  & & & 3.78 \% & \\  

                \hline \hline 
                & Total & & & 8,224  & & & 67.87 \% & \\  
                \hline
            \end{tabular}
            \label{table:selected-six-sensors}
    \end{table}

When training the DCRNN model for various local locations, the strategy is to begin by examining two locations and then gradually increase the number of new locations. This allows us to assess whether or not the model improves with multiple sensors. Starting off with two sensor locations, the DCRNN obtained an MAE value of 16.25 which is an acceptable score, however when onboarding two additional sensors, the MAE value resulted in 14.34, with the final MAE score of 11.8 when onboarding all six sensors. Additionally, a comparison of different batch sizes (10 vs 50) is also tested to check which of both would perform better and if there would be a noticeable difference, if any, resulting that a batch size of 50 was more efficient. Throughout this section, we analyzed the data collected through the models produced in the previous section, thus addressing Objective \textbf{O3}.

\section{Conclusion}

The research aimed to develop a traffic prediction model for local traffic management. The study compared different traffic flow datasets from various cities globally and introduced a new dataset called MalTra (\textbf{O1}). The research trained and analyzed a traffic prediction model using a large-scale dataset and evaluated its relevance at the local level (\textbf{O2}). The study found that the DCRNN model outperformed the STGCN model in terms of MAE, MAPE, and RMSE for both datasets (\textbf{O3}). The findings suggest that traffic prediction in Malta is possible with a more detailed dataset.

\end{document}